%% file: main_singleblind.tex
\documentclass{article}
\usepackage{spconf,amsmath,graphicx}
\usepackage{epsfig}
\usepackage{makecell}
\usepackage{amsfonts}
\usepackage{hyperref}
\usepackage{xcolor}
\usepackage{subcaption}
\usepackage{multirow}
\usepackage{amssymb}
\usepackage[numbers]{natbib}
\usepackage{booktabs}
\usepackage{color, colortbl}
\definecolor{Gray}{RGB}{255, 200, 220}
\usepackage{enumitem}

\definecolor{mygray}{gray}{0.9}

\let\OLDthebibliography\thebibliography
\renewcommand\thebibliography[1]{
  \OLDthebibliography{#1}
  \setlength{\parskip}{0pt}
  \setlength{\itemsep}{0pt plus 0.3ex}
}

\title{Information Router for Mitigating Modality Dominance in Vision-Language Models}
\name{Seulgi Kim, Mohit Prabhushankar, Ghassan AlRegib\thanks{This work is supported by the ML4Seismic Consortium at Georgia Tech.}}

\address {OLIVES at the Center for Signal and Information Processing CSIP,\\ 
School of Electrical and Computer Engineering, Georgia Institute of Technology, Atlanta, GA, USA \\
\{seulgi.kim, mohit.p, alregib\}@gatech.edu             }

\begin{document}









\sloppy
%
\maketitle
\input{Sections/01-abstract}
\input{Sections/02-intro}
\input{Sections/03-related-works}

\input{Sections/05-methodology}

\input{Sections/06-experiments}

\input{Sections/07-conclusion}

\bibliographystyle{IEEEbib}
\bibliography{strings,refs}

\end{document}

%% file: Sections/01-abstract.tex
\begin{abstract}
Vision Language models (VLMs) have demonstrated strong performance across a wide range of benchmarks, yet they often suffer from modality dominance, where predictions rely disproportionately on a single modality. Prior approaches primarily address this issue by steering model's attention allocation, implicitly assuming that all modalities provide sufficient information. However, attention only determines where the model focuses, and cannot enrich information that is missing or ambiguous. In the real world, input modalities often differ in information density and their signal-to-noise ratios. In such cases, simply adjusting model's attention does not resolve the underlying lack of information. In this paper, we propose \textsc{MoIR}: \textit{Multi-modal Information Router}, an information-level fusion method that explicitly reduces information disparity prior to fusion. \textsc{MoIR} identifies less informative tokens and routes complementary information from a stronger modality, constructing information-dense token representations before they are processed by a large language model. By modifying information availability, \textsc{MoIR} enables reliable shifts in modality dominance, even when one modality is degraded. We evaluate \textsc{MoIR} on three widely used multi-modal benchmarks across multiple model backbones. Experimental results show that \textsc{MoIR} consistently demonstrates more balanced modality contribution, and improves robustness and downstream performance, particularly even under modality degradation. These findings demonstrate that explicitly modifying cross-modal information is an effective and complementary strategy for mitigating modality dominance in multi-modal reasoning models.
\end{abstract}


%% file: Sections/02-intro.tex
\begin{figure}[ht!]
\begin{center}
\includegraphics[width=\linewidth]{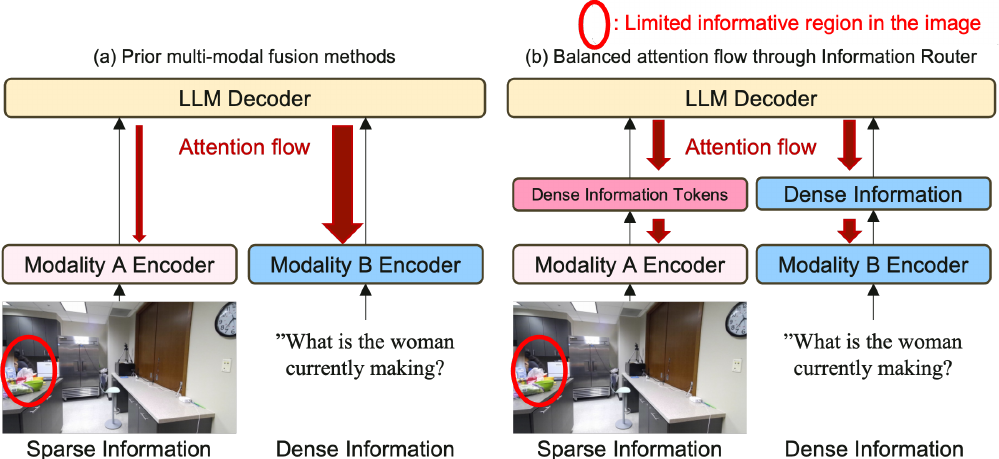}
\end{center}
\vspace{-0.3cm}
   \caption{This figure compares between existing Vision-Language model architecture and \textsc{MoIR}. Red arrows indicate attention flow from the LLM decoder to modality-specific tokens. The width of each arrow represents the average attention magnitude per token. (a) Modality-specific tokens are directly fed into the LLM decoder, without accounting for cross-modal information disparity. As a result, differences in information density across modalities lead to imbalanced attention allocation and modality dominance. (b) \textsc{MoIR} selectively routes complementary information from the other modality to construct information-dense tokens. By reducing token-level information disparity, \textsc{MoIR} enables more balanced attention flow across modalities. }
\label{fig:comparison}
\end{figure}

\section{Introduction}
\label{sec:Intro}

Vision-Language Models (VLMs) have achieved strong performance across a wide range of benchmarks~\citep{chen2024internvl, kim2021vilt, sim2025can,wu2023multimodal,peng2024synthesize, cai2025mllm, lu2022learn,liu2024improved,bai2025qwen2,hu2022lora, gurari2018vizwiz, fang2024mmbench}. Despite this progress, modality dominance, where models often rely disproportionately on a single modality, remains a persistent challenge~\citep{sim2025can,cai2025mllm,wu2025language,javaloy2022mitigating,park2025assessing}. Especially, vision-language models may produce correct answers while paying little attention to visual evidence~\citep{sim2025can,wu2025language}, raising the question of whether improved accuracy reflects genuine multi-modal reasoning or uni-modal shortcuts. This can cause performance failure, especially when one modality is corrupted and a complementary modality is needed~\citep{sim2025can,cai2025mllm,wu2025language,javaloy2022mitigating,park2025assessing}. As a result, understanding and mitigating modality dominance is essential for building reliable multi-modal systems~\citep{sim2025can,cai2025mllm,wu2025language,javaloy2022mitigating,park2025assessing}. 

Recent works have proposed various methods to mitigate modality dominance~\citep{sim2025can,cai2025mllm,wu2025language,javaloy2022mitigating,park2025assessing}. As shown in Figure~\ref{fig:comparison} (a), these approaches diagnose modality dominance through the magnitude of attention weights, and mitigate the modality dominance by steering the model's attention allocation~\citep{wu2025language}. While effective, these approaches implicitly assume that both input modalities already contain sufficient information, and that dominance arises mainly from the attention imbalance. However, in the real world, modalities often differ substantially in information density. For example, as shown in Figure~\ref{fig:comparison} (a-b), visual tokens may be sparse or ambiguous due to exocentric view or viewpoint limitations, while language tokens tend to be dense and semantically explicit. When such heterogeneous tokens are jointly processed without accounting for this information disparity, information imbalance across modalities can manifest as modality dominance.

To counter this information disparity, we propose \textsc{MoIR}: \textit{Multi-modal Information Router}, a multi-modal fusion method that explicitly routes information prior to fusion. Here, we define a token to be informative if its embedding contributes strongly to the dominant directions of the representation space, following the intuition in~\citep{kokilepersaud2024hex,kim2025countering, kokilepersaud2025adadim}. As illustrated in Figure~\ref{fig:comparison} (b) and Figure~\ref{fig:architecture}, \textsc{MoIR} selectively routes complementary information across modalities to construct information-dense tokens, thereby reducing information disparity between modalities before fusion. In addition, we show that such information routing leads to lower Attention Efficiency Index (AEI) values, indicating more balanced modality usage. To assess the effectiveness of \textsc{MoIR}, we conduct comprehensive experiments on three widely used datasets and model backbones. Our key contributions are:
\vspace{-0.2cm}
\begin{enumerate}
    \item We propose \textsc{MoIR}, a multi-modal fusion method with a cross-modal information exchange, which addresses the modality dominance problems.
    \item We revisit the definition of modality dominance and show that changing the modality usage without considering the evidential status of each modality can be unreliable when a modality is noisy or uninformative.
    \item We provide extensive empirical results showing that \textsc{MoIR} consistently changes modality contribution and improves robustness and/or downstream performance across diverse datasets and models.
\end{enumerate}

%% file: Sections/03-related-works.tex
\section{Related Work}
\vspace{-0.2cm}
\subsection{Multi-modal Fusion}
Multi-modal learning aims to integrate heterogeneous data sources to achieve performance superior to unimodal approaches~\citep{akkus2023multimodal, chen2024internvl, kim2021vilt}. Fusion strategies are broadly categorized into three strategies. First, aggregation-based fusion strategies combine features through concatenations, summation, or attention pooling~\citep{kaviani2025hierarchical}. Second, alignment-based fusion methods temporally or semantically align modality representations~\citep{kim2025multi, kim2021vilt, kim2025countering}. Finally, hybrid approaches integrate both aggregation and alignment techniques~\citep{baltruvsaitis2018multimodal}. Despite their architectural advances, these fusion strategies often suffer from feature redundancy, where redundant or less informative features from one modality can overwhelm the shared representation space, resulting in suboptimal fusion~\citep{kim2025countering}. Our method addresses this limitation by selectively routing complementary information across modalities, improving the informativeness of token representations for both modalities.

\vspace{-0.2cm}
\subsection{Modality Dominance}
\vspace{-0.3cm}
Multi-modal models frequently exhibit a bias toward specific modalities while neglecting others. This phenomenon has been extensively discussed in the literature under various definitions, including modality dominance~\citep{wu2025language}, modality collapse~\citep{sim2025can, javaloy2022mitigating}, and the greedy nature of learning~\citep{sim2025can, wu2023multimodal}. First, a significant body of research predominantly focuses on rectifying the learning trajectory by manipulating loss functions or gradient flows during training. Specifically, ~\citep{wu2023multimodal, javaloy2022mitigating, peng2024synthesize, zhou2023intra} posits that models inherently prioritize modalities that are easier to learn, which causes the optimization or gradient flow of other modalities to stagnate prematurely. Other studies have focused on internal model mechanics. ~\citep{wu2025language} argues that non-text modalities suffer from token redundancy compared to the high information density of text. This leads the model to disproportionately allocate attention to text tokens. Also, ~\citep{park2025assessing, cai2025mllm} highlight the data's intrinsic characteristics, such as modality-agnostic questions that can be answered using text priors alone or lack of semantic completeness in video-query pairs. While these approaches mitigate modality dominance by changing model architecture or training dynamics, they often overlook the fundamental characteristics of the input data itself. These methods implicitly assume that the issue lies solely in the model's processing, neglecting the information disparity inherent in the input tokens prior to fusion. In contrast, \textsc{MoIR} addresses the root cause by explicitly routing token information density before fusion.

%% file: Sections/05-methodology.tex
\begin{figure}[t!]
\begin{center}
\includegraphics[width=\linewidth]{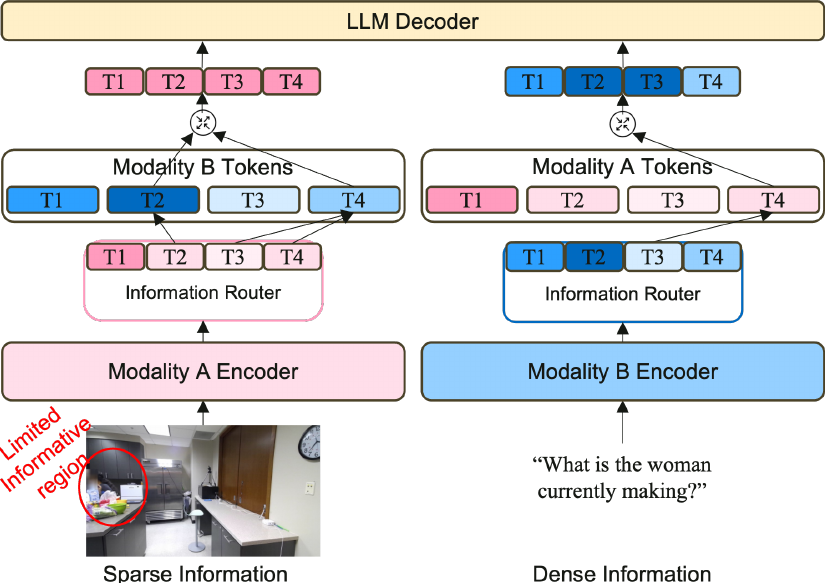}
\end{center}
\vspace{-0.2cm}
   \caption{Overview of \textsc{MoIR}. \textcolor{magenta}{Modality A} and \textcolor{blue}{Modality B} are first encoded into token sequences $T1, T2, T3, T4$. The color intensity of each token represents its informativeness. For example, lighter color indicates sparse and less informative tokens and darker colors indicate dense and more informative tokens. \textsc{MoIR} identifies less informative channels within tokens and adaptively routes information from a complementary modality before tokens are processed by the LLM decoder.}
\label{fig:architecture}
\vspace{-0.5cm}
\end{figure}

\section{Methodology}
\label{sec:methodology}
This section presents \textsc{MoIR}: \textit{Multi-modal Information Router}, which controls the information flow between modalities. As illustrated in Figure~\ref{fig:architecture}, \textsc{MoIR} operates between modality-specific encoders and the LLM decoder, where modality dominance often manifests as imbalanced information utilization. Here, unlike previous methods that enforce balanced attention, \textsc{MoIR} directly modifies token representations before fusion by routing complementary information into less informative channels.

\vspace{-0.15cm}
\subsection{Multi-modal Token Representations}
\vspace{-0.1cm}
Given a multi-modal input consisting of two modalities, we first obtain modality-specific token representations using their respective encoders, as denoted in Figure~\ref{fig:architecture} as $T1, T2, T3,T4$. Let $\mathbf{F}^{A} \in \mathbb{R}^{B \times L_A \times D},
\mathbf{F}^{B} \in \mathbb{R}^{B \times L_B \times D}$ denote the token sequences of modality~$A$ and modality~$B$, respectively, where $B$ is the batch size, $L_A$ and $L_B$ are the numbers of tokens produced by each encoder, and $D$ is the embedding dimension used by the LLM. Then, each token $t$ from modality $A$ is given by $\mathbf{f}^{A}_t \in \mathbb{R}^{D}$, whose dimensions are referred to as channels. These tokens encode modality-specific information and serve as inputs to the information router prior to fusion by the LLM decoder.

\vspace{-0.15cm}
\subsection{Identifying Less informative Tokens}
\vspace{-0.1cm}
Not all tokens contribute equally to multi-modal reasoning. In practice, a token can be ineffective, or less informative, when its representation is dominated by weak, noisy, or redundant channels. \textsc{MoIR} identifies these less informative channels as the underlying cause of less informative tokens, and uses them to determine where information routing is required. For modality~$A$, we reshape the token sequence
\(
\mathbf{F}^{A} \in \mathbb{R}^{B \times L_A \times D}
\)
into
\(
\tilde{X}^{A} \in \mathbb{R}^{(B \cdot L_A) \times D}
\),
and perform singular value decomposition:
$\tilde{X}^{A} = U^{A} \Sigma^{A} (V^{A})^\top$, where $\Sigma^{A} = \mathrm{diag}(\sigma^{A}_1, \dots, \sigma^{A}_r)$ and $V^{A} = [v^{A}_1, \dots, v^{A}_r]$. We define the informativeness score for channel $d$ as
\begin{equation}
S^{A}_d = \sum_{i=1}^{r} (\sigma^{A}_i)^2 (v^{A}_{i,d})^2 .
\end{equation}
This measures contribution of each channel to the principal components of the token representation. Channels with low $S^{A}_d$ are considered less informative, and tokens dominated by such channels provide limited information. The same procedure is applied to modality~$B$ to obtain $S^{B}_d$. For each modality, we select the bottom $k'$ channels with the lowest informativeness scores as candidates for information routing:
\begin{equation}
\mathcal{D}^{A}_{\text{less informative}} = \text{Bottom}_{k'} S^{A}_d, \quad
\mathcal{D}^{B}_{\text{less informative}} = \text{Bottom}_{k'} S^{B}_d .
\end{equation}

\vspace{-0.15cm}
\subsection{Multi-Modal Information Routing}
\vspace{-0.1cm}
Once less-informative channels are identified, as shown in Figure~\ref{fig:architecture}, \textsc{MoIR} selectively modifies token representations by routing complementary information to compensate for insufficient information along these channels. Specifically, for modality~$A$, the routed token sequence is computed as
\begin{equation}
\bar{\mathbf{F}}^{A}_{:,:, \mathcal{D}^{A}_{\text{less informative}}}
=
\alpha_{\mathcal{D}^{A}} \cdot
\mathbf{F}^{B}_{:,:, \mathcal{D}^{A}_{\text{less informative}}}
+
\left(1 - \alpha_{\mathcal{D}^{A}}\right) \cdot
\mathbf{F}^{A}_{:,:, \mathcal{D}^{A}_{\text{less informative}}},
\end{equation}
with the same formulation applied to modality~$B$. Here, $\alpha_{\mathcal{D}^{A}}, \alpha_{\mathcal{D}^{B}}$ are learnable routing gates that control the amount of information injected into each channel. This routing mechanism allows \textsc{MoIR} to directly modify the information carried by tokens before fusion, enhancing the informativeness of less informative tokens while preserving already informative components.

\vspace{-0.2cm}
\subsection{Integration with the LLM Decoder}
\vspace{-0.1cm}
As shown on the top of Figure~\ref{fig:architecture}, the routed token sequences $\bar{\mathbf{F}}^{A}$ and $\bar{\mathbf{F}}^{B}$ are concatenated and passed to the LLM decoder. By routing complementary information into less informative channels, \textsc{MoIR} enables the LLM to naturally adjust its reliance across modalities based on the available information. In Section~\ref{sec:experiments}, we show that this information routing also leads to increased Attention Efficiency Index (AEI) values and more balanced modality contribution, indicating that changes in modality usage emerge as a consequence of improved information availability.

%% file: Sections/06-experiments.tex
\input{table/main}
\section{Experiments}
\label{sec:experiments}
\vspace{-0.2cm}
In this section, we evaluate the effectiveness of \textsc{MoIR} in mitigating modality dominance across diverse multi-modal settings. Our experiments are designed to answer the following questions:  
(1) Does \textsc{MoIR} mutually improve informativeness for both modalities?
(2) Does \textsc{MoIR} improve modality balance without degrading task performance?
(3) Is \textsc{MoIR} robust under modality degradation or bias?

\vspace{-0.2cm}
\subsection{Experimental Setup}
\noindent \textbf{Datasets.} We aim to show that \textsc{MoIR} can be applied to different forms of modality dominance, ranging from text-dominant to video-dominant settings. Hence, we evaluate \textsc{MoIR} on three representative multi-modal benchmarks covering different dominance regimes: ScienceQA~\citep{lu2022learn} and VizWiz~\citep{gurari2018vizwiz} for image-text reasoning, and MMBench-Video~\citep{fang2024mmbench} for video-audio-text reasoning.

\vspace{0.1cm}
\noindent \textbf{Implementation Details.} We fine-tune LLaVA-1.5-7B~\citep{liu2024improved}, LLaVA-1.5-13B~\citep{liu2024improved} and Qwen2.5-VL~\citep{bai2025qwen2}. Since fine-tuning all VLM parameters can be computationally inefficient and prone to catastrophic forgetting~\citep{kirkpatrick2017overcoming}, we adopt parameter-efficient fine-tuning using LoRA~\citep{hu2022lora}. During fine-tuning, \textsc{MoIR} is inserted between modality-specific encoders and the LLM decoder without altering the underlying model architecture. Specifically for LLaVA-1.5-7B and LLaVA-1.5-13B~\citep{liu2024improved}, \textsc{MoIR} is applied right before the image embeddings are interleaved into the text embedding sequence. For Qwen2.5-VL~\citep{bai2025qwen2}, \textsc{MoIR} is applied after the vision encoder produces visual tokens and before LLM decoder.

\vspace{0.1cm}
\noindent \textbf{Hyperparameter Details.} All models are trained for 10 epochs following~\citep{lu2022learn, gurari2018vizwiz}, with AdamW~\citep{loshchilov2017decoupled} optimizer, a learning rate of $2\times10^{-4}$ and no weight decay, a batch size of 8, with a plateau-based learning rate scheduler. For parameter-efficient fine-tuning, we apply LoRA~\citep{hu2022lora} with rank $r=16$, scaling factor $\alpha=32$, and dropout rate $0.05$. The maximum input sequence length is set to 2048 tokens. For \textsc{MoIR}, the routing coefficients $\alpha_{\mathcal{D}^{A}}$ and $\alpha_{\mathcal{D}^{B}}$ are initialized to $0.5$. We set the exchange ratio $k'$ to $0.10$, but we conduct ablation studies by varying the routing ratio applied to the lowest-importance channels to analyze the sensitivity of \textsc{MoIR} to the degree of information routing. All experiments are conducted on a single NVIDIA H100-SXM5-80GB GPU. 

\vspace{0.1cm}
\noindent \textbf{Evaluation Setup.} For downstream task performance, we follow the official benchmark protocol written in the original dataset benchmark papers~\citep{lu2022learn, gurari2018vizwiz, fang2024mmbench}.
To quantify modality usage, we report the Modality Dominance Index (MDI) and Attention Efficiency Index (AEI) following prior work~\citep{wu2025language}. To analyze the changes in representation informativeness with \textsc{MoIR}, we measure the effective rank of token representation matrices following prior work~\citep{kim2025countering}. This measurement shows the distribution of representations across singular value directions~\citep{roy2007effective}.

\vspace{-0.2cm}
\subsection{Quantitative Results}
\noindent \textbf{Downstream Performance.} The column \textbf{Acc.} in Table~\ref{tab:main}  shows the comparison of downstream task performance between standard fine-tuning (FT) and FT with \textsc{MoIR}. Across all benchmarks, \textsc{MoIR} maintains or improves task accuracy while significantly reducing modality dominance. Specifically, on ScienceQA, the accuracy gains from \textsc{MoIR} are relatively modest. We attribute this to the multiple-choice form of the dataset, where the answer is constrained and even partially grounded reasoning can still lead to correct answer selection. In contrast, clearer performance improvements are observed in VizWiz and MMBench-Video, which require stronger visual grounding and open-ended reasoning. Especially, the most significant gains appear on MMBench-Video, that requires temporal reasoning over dynamic visual content. These improvements indicate that \textsc{MoIR} is effective in scenarios where visual evidence is indispensable and cannot be substituted by textual inputs.

\vspace{0.1cm}
\noindent \textbf{Modality Dominance Analysis.} The column \textbf{Information} in Table~\ref{tab:main} shows information density, and the column \textbf{MDI, AEI} in Table~\ref{tab:main} shows modality dominance reduction rate on both image-text benchmarks and video-audio-text benchmarks. Across all datasets and placements, \textsc{MoIR} consistently increases Rank $\Delta$ I and Rank $\Delta$ T. This increase in both image and text suggests that \textsc{MoIR} does not merely suppress one modality but enhances complementary information exchange, resulting in more information-dense representations. Also, \textsc{MoIR} substantially reduces MDI and AEI across nearly all configurations. This result shows that \textsc{MoIR} mitigates a disproportionate reliance on a single modality.

\input{table/image-swapping}
\vspace{0.1cm}
\noindent \textbf{Robustness.} We further analyze modality reliance under visual corruption. As shown in Table~\ref{tab:image_replacement}, the baseline model produces the same predictions (62.13\%) even after the image is replaced with noise, suggesting a substantial reliance on language even when visual input is removed. In particular, even for vision-dependent questions that require visual evidence, the model still maintains a high unchanged rate (57.80\%). In contrast, \textsc{MoIR} significantly reduces the rate, indicating that predictions are more sensitive to visual inputs. These results demonstrate that \textsc{MoIR} mitigates spurious language reliance and promotes evidence-based multi-modal reasoning.

\input{table/ablation}
\vspace{0.1cm}
\noindent \textbf{Ablation Study.} 
As shown in Table~\ref{tab:ablation}, we vary the exchange ratio $k'$ on  VizWiz dataset with LoRA applied to attention layers of LLaVA-1.5-7B model. We observe that $k'=0.1$ achieves the best downstream performance and strongest information enhancement (Rank $\Delta$I, Rank $\Delta$T), while also maintaining competitive modality balance (MDI, AEI). 

\begin{figure}[t!]
\begin{center}
\includegraphics[width=1.02\linewidth]{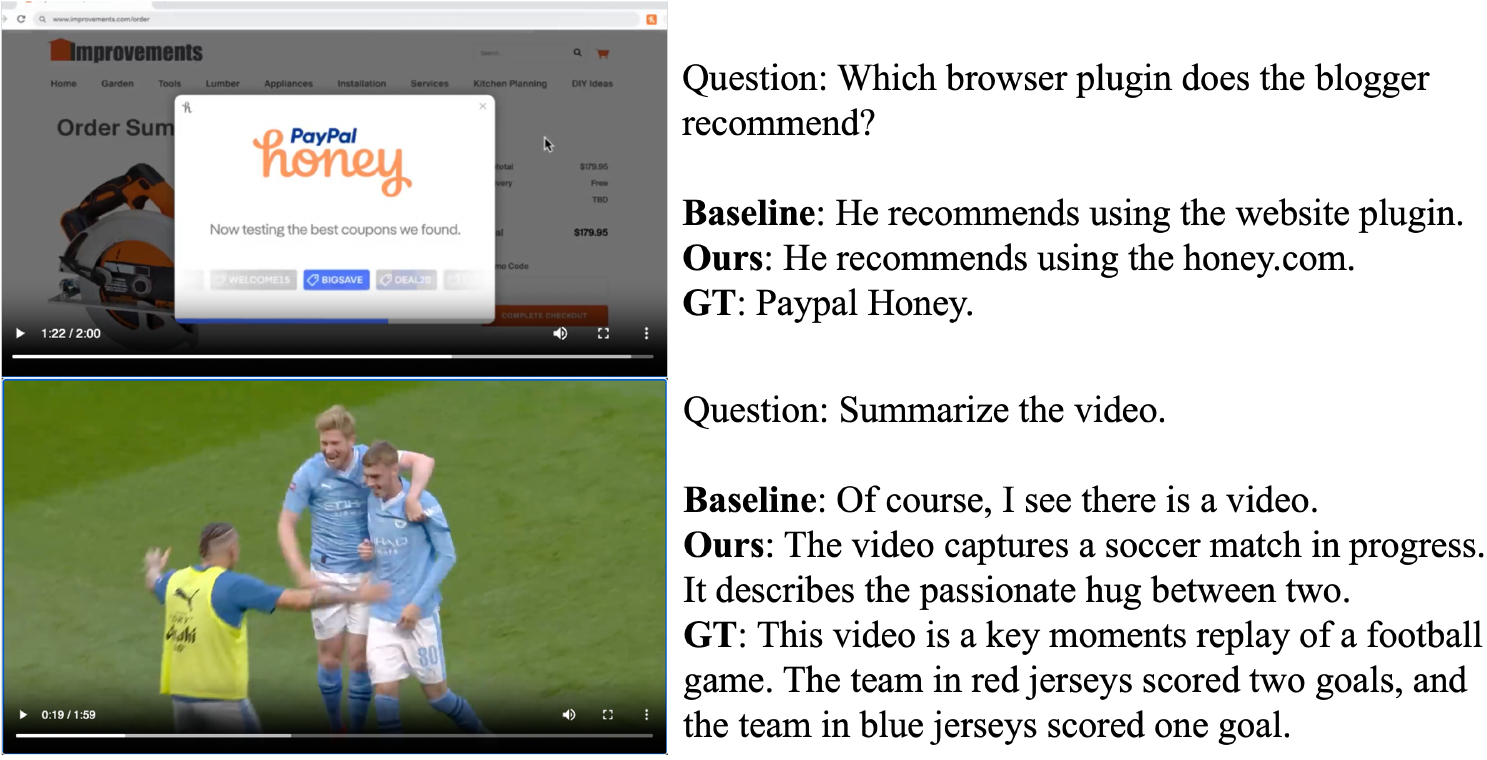}
\end{center}
\vspace{-0.4cm}
   \caption{This shows the qualitative comparison on MMBench-Video dataset. The baseline often generates responses that either mimic the wording in the question or produce hallucinated answers that are not grounded in the video content. In contrast, \textsc{MoIR} produces video-grounded responses by relying on visual evidence from the frames.}
\label{fig:qualitative}
\vspace{-0.2cm}
\end{figure}

\subsection{Qualitative Results}
Figure~\ref{fig:qualitative} presents qualitative comparisons on the MMBench-Video dataset. The baseline (standard fine-tuning) model frequently generates responses that either mimic the wording of the question (`website plugin'), or produce generic statements (`I see there is a video'), that are not grounded in the video content. In contrast, as shown in Figure~\ref{fig:qualitative}, it correctly identifies `Paypal Honey' as the plugin, and provides a more accurate description of the soccer match by referencing observable events. These examples further demonstrate that \textsc{MoIR} reduces spurious language reliance and encourages video-grounded reasoning, leading to more faithful and visually consistent predictions.

%% file: table/main.tex

\begin{table*}[t]
\centering
\setlength{\tabcolsep}{4pt}
\renewcommand{\arraystretch}{1.12}
\scriptsize
\resizebox{0.8\textwidth}{!}{
\begin{tabular}{l c c l c c cc cc}
\toprule
Dataset & Modality & Model & Place. & Training & Acc. ($\uparrow$) &
\multicolumn{2}{c}{Information} & MDI ($\downarrow$) & AEI ($\downarrow$) \\
\cmidrule(lr){7-8}
 &  &  &  &  &  & Rank $\Delta$I ($\uparrow$) & Rank $\Delta$T ($\uparrow$) &  & \\
\midrule

\multirow{6}{*}{ScienceQA~\cite{lu2022learn}} &
\multirow{6}{*}{I+T} &
\multirow{6}{*}{LLaVA-1.5-7B~\cite{liu2024improved}} &
Attn & FT & \textbf{73.68} & 12.35 & 30.98 & 198.49 & 8.70 \\
& & & & \cellcolor{Gray} FT w/ \textsc{MoIR}  & \cellcolor{Gray} 70.53 & \cellcolor{Gray}\textbf{35.95} & \cellcolor{Gray}\textbf{35.07} & \cellcolor{Gray} \textbf{89.97} &\cellcolor{Gray} \textbf{8.21} \\
\cmidrule(lr){4-10}
& & & MLP & FT  & \textbf{71.58} & 12.37 & 30.64 & 192.72 & 8.69 \\
& & & & \cellcolor{Gray}FT w/ \textsc{MoIR}  & \cellcolor{Gray}70.53 & \cellcolor{Gray}\textbf{32.59} & \cellcolor{Gray}\textbf{33.73} &  \cellcolor{Gray}\textbf{79.83} & \cellcolor{Gray}\textbf{8.12} \\
\cmidrule(lr){4-10}
& & & Proj. & FT& 62.11 & 12.39 & 29.90 & 192.11 & 8.69 \\
& & & & \cellcolor{Gray}FT w/ \textsc{MoIR} & \cellcolor{Gray}\textbf{67.37} & \cellcolor{Gray}\textbf{35.68} & \cellcolor{Gray}\textbf{32.85} &  \cellcolor{Gray}\textbf{88.30} & \cellcolor{Gray}\textbf{8.19} \\
\midrule

\multirow{6}{*}{ScienceQA~\cite{lu2022learn}} &
\multirow{6}{*}{I+T} &
\multirow{6}{*}{LLaVA-1.5-13B~\cite{liu2024improved}} &
Attn & FT & 83.16 & 66.34 & 35.44 & 100.03 & 8.27 \\
& & & & \cellcolor{Gray} FT w/ \textsc{MoIR}  & \cellcolor{Gray} \textbf{83.42} & \cellcolor{Gray}\textbf{71.25} & \cellcolor{Gray}\textbf{36.30} & \cellcolor{Gray} \textbf{91.58} &\cellcolor{Gray} \textbf{8.22} \\
\cmidrule(lr){4-10}
& & & MLP & FT  & 66.32 & \textbf{65.96} & 34.82 & 96.39 & 8.25 \\
& & & & \cellcolor{Gray}FT w/ \textsc{MoIR}  & \cellcolor{Gray}\textbf{69.47} & \cellcolor{Gray}65.47 & \cellcolor{Gray}\textbf{34.94} &  \cellcolor{Gray}\textbf{94.62} & \cellcolor{Gray}\textbf{8.24} \\
\cmidrule(lr){4-10}
& & & Proj. & FT& \textbf{65.26} & \textbf{52.23} & 36.22 & 79.09 & 8.12 \\
& & & & \cellcolor{Gray}FT w/ \textsc{MoIR} & \cellcolor{Gray}63.15 & \cellcolor{Gray}47.23 & \cellcolor{Gray}\textbf{37.25} &  \cellcolor{Gray}\textbf{73.15} & \cellcolor{Gray}\textbf{8.06} \\
\midrule

\multirow{6}{*}{VizWiz~\cite{gurari2018vizwiz}} &
\multirow{6}{*}{I+T} &
\multirow{6}{*}{LLaVA-1.5-7B~\cite{liu2024improved}} &
Attn & FT & 28.22 & 18.32 & 31.83 &  81.70 & 10.97 \\
& & & & \cellcolor{Gray}FT w/ \textsc{MoIR} & \cellcolor{Gray}\textbf{32.47} & \cellcolor{Gray}\textbf{20.65} & \cellcolor{Gray}\textbf{32.88} &  \cellcolor{Gray}\textbf{73.16} & \cellcolor{Gray}\textbf{10.43} \\
\cmidrule(lr){4-10}
& & & MLP & FT  & \textbf{33.38} & \textbf{19.87} & \textbf{29.75} &  91.19 & 10.72 \\
& & & & \cellcolor{Gray}FT w/ \textsc{MoIR}  & \cellcolor{Gray} 30.16 & \cellcolor{Gray}17.95 & \cellcolor{Gray}26.39 &  \cellcolor{Gray}\textbf{84.73} & \cellcolor{Gray}\textbf{10.65} \\
\cmidrule(lr){4-10}
& & & Proj. & FT& 43.50 & 16.86 & 27.25 & 125.38 & 11.32 \\
& & & & \cellcolor{Gray}FT w/ \textsc{MoIR} & \cellcolor{Gray}\textbf{47.00} & \cellcolor{Gray}\textbf{16.89} & \cellcolor{Gray}\textbf{27.39} & \cellcolor{Gray}\textbf{118.19} & \cellcolor{Gray}\textbf{10.93} \\
\midrule

\multirow{6}{*}{MMBench-Video~\cite{fang2024mmbench}} &
\multirow{6}{*}{V+A+T} &
\multirow{6}{*}{Qwen2.5-VL~\cite{bai2025qwen2}} &
Attn & FT & 69.57 & 445.05 & 11.60 & 461.23 & 30.80 \\
& & & & \cellcolor{Gray}FT w/ \textsc{MoIR} & \cellcolor{Gray}\textbf{71.43} & \cellcolor{Gray}\textbf{489.42} & \cellcolor{Gray}\textbf{14.46} & \cellcolor{Gray}\textbf{461.00} & \cellcolor{Gray}\textbf{30.74} \\
\cmidrule(lr){4-10}
& & & MLP & FT  & 42.86 & 651.26 & 27.62 & \textbf{296.46} & \textbf{29.62} \\
& & & & \cellcolor{Gray}FT w/ \textsc{MoIR}  & \cellcolor{Gray}\textbf{100.00} & \cellcolor{Gray}\textbf{685.06} & \cellcolor{Gray}\textbf{29.62} & \cellcolor{Gray}374.95 & \cellcolor{Gray}30.82 \\
\cmidrule(lr){4-10}
& & & Proj. & FT& 52.38 & 651.46 & 27.62 & \textbf{296.96} & \textbf{29.49} \\
& & & & \cellcolor{Gray}FT w/ \textsc{MoIR}& \cellcolor{Gray}\textbf{60.87} & \cellcolor{Gray}\textbf{1929.29} & \cellcolor{Gray}\textbf{28.25} & \cellcolor{Gray}328.41 & \cellcolor{Gray}39.08 \\
\bottomrule
\end{tabular}}
\caption{This table summarizes performance, representational changes, and modality usage across diverse configurations. \textbf{Modality} indicates the input modalities (\textbf{I+T}: Image + Text, \textbf{V+A+T}: Video + Audio + Text). \textbf{Place.} denotes the layer type to which LoRA is applied: \textbf{Attn} denotes attention layers, \textbf{MLP} denotes feed-forward layers, and \textbf{Proj.} is a projection layers. \textbf{Acc.} reports the official benchmark accuracy from ~\citep{gurari2018vizwiz, fang2024mmbench, lu2022learn}. \textbf{Rank$\Delta$I} and \textbf{Rank$\Delta$T} denote changes in the effective rank of image/video and text token representation spaces, respectively. \textbf{MDI} denotes Modality Dominance Index~\citep{wu2025language} and \textbf{AEI} denotes Attention Efficiency Index~\citep{wu2025language}. Results are shown for standard LoRA fine-tuning (\textbf{LoRA FT}) and LoRA fine-tuning with the proposed Multi-modal Information Router (\textbf{FT w/ \textsc{MoIR}}). Higher Rank $\Delta$I / Rank $\Delta$T indicates stronger information contribution (higher value is better (denoted as $\uparrow$))~\citep{kim2025countering}, whereas lower MDI and AEI indicate more balanced attention usage (lower value is better (denoted as $\downarrow$).~\citep{wu2025language}}
\label{tab:main}
\vspace{-0.3cm}
\end{table*}

%% file: table/image-swapping.tex




\newcolumntype{C}[1]{>{\centering\arraybackslash}p{#1}}

\begin{table}[t]
\centering
\scriptsize
\setlength{\tabcolsep}{3pt}
\renewcommand{\arraystretch}{1.15}

\begin{tabular}{lccc}
\toprule
& \textbf{All }
& \textbf{Vision-dependent questions}
& \textbf{Vision-irrelevant questions} \\
\midrule
FT & 62.13\% & 57.80\% & 69.33\% \\
\rowcolor{Gray} FT + \textsc{MoIR}     & 29.63\% & 25.20\% & 37.00\% \\
\bottomrule
\end{tabular}
\caption{\textbf{Robustness.} We replace each original image in the VizWiz dataset with Gaussian noise, and measure the rate of unchanged model predictions. \textbf{Vision-dependent questions} refer to questions whose correct answers require visual evidence from the image, whereas \textbf{Vision-irrelevant questions} can be answered without relying on visual content. A high unchanged rate means the model is insensitive to visual input.}
\label{tab:image_replacement}
\end{table}

%% file: table/ablation.tex

\begin{table}[t]
\centering
\small
\setlength{\tabcolsep}{5pt}
\renewcommand{\arraystretch}{1.05}

\resizebox{\linewidth}{!}{
\begin{tabular}{lccccc}
\toprule
$k'$ & Acc. ($\uparrow$) & \multicolumn{2}{c}{Information} & MDI ($\downarrow$) & AEI ($\downarrow$) \\
\cmidrule(lr){3-4}
 &  &  Rank $\Delta I$ ($\uparrow$) & Rank $\Delta T$ ($\uparrow$) &  &  \\
\midrule
0.05 & 31.53 & 18.94 & 32.64 & \textbf{69.97} & 10.42 \\
0.10 & \textbf{32.47} & \textbf{20.65} & \textbf{32.88} & 73.16 & 10.43 \\
0.15 & 30.06 & 20.42 & 31.60 & 79.97 & \textbf{10.22} \\
\bottomrule
\end{tabular}
}
\vspace{-0.2cm}
\caption{This table shows the ablation studies by varying the exchange ratio $k'$ in VizWiz dataset, while LoRA applied to the attention layer of LLaVA-1.5-7B model.}
\label{tab:ablation}
\vspace{-0.2cm}
\end{table}

%% file: Sections/07-conclusion.tex
\section{Conclusion}
\vspace{-0.2cm}
In this paper, we revisit the problem of modality dominance in vision–language models from an information-centric perspective. Prior approaches assume that all modalities already provide sufficient information, which could not hold in realistic settings. To address this limitation, we propose \textsc{MoIR}, a Multi-modal Information Router that reduces information disparity across modalities. Concretely, \textsc{MoIR} identifies less informative tokens and selectively routes complementary information from a stronger modality to construct information-dense token representations. Extensive experiments across multiple benchmarks and model backbones demonstrate that \textsc{MoIR} consistently leads to more balanced modality contribution and improves robustness and downstream performance. This suggests that manipulating information is an effective and complementary strategy for mitigating modality dominance, and highlights the importance of information-level interventions for reliable multi-modal reasoning.